\documentclass[10pt,twocolumn,letterpaper]{article}

\usepackage{3dv}
\usepackage{times}
\usepackage{epsfig}
\usepackage{graphicx}
\usepackage{amssymb}
\usepackage{subfigure}
\usepackage{color}
\usepackage{float}

\threedvfinalcopy 

\def\threedvPaperID{****} 

\ifthreedvfinal\fi
\begin{document}

\title{Performance Evaluation of 3D Correspondence Grouping Algorithms}

\author{Jiaqi Yang, Ke Xian, Yang Xiao and Zhiguo Cao\\
	School of Automation, Huazhong University of Science and Technology\\
	Wuhan, P. R. China\\
	{\tt\small \{jqyang, kexian, Yang\_Xiao, zgcao\}@hust.edu.cn}
}

\maketitle

\begin{abstract}
	This paper presents a thorough evaluation of several widely-used 3D correspondence grouping algorithms, motived by their significance in vision tasks relying on correct feature correspondences. A good correspondence grouping algorithm is desired to retrieve as many as inliers from initial feature matches, giving a rise in both precision and recall. Towards this rule, we deploy the experiments on three benchmarks respectively addressing shape retrieval, 3D object recognition and point cloud registration scenarios. The variety in application context brings  a rich category of nuisances including noise, varying point densities, clutter, occlusion and partial overlaps. It also results to different ratios of inliers and correspondence distributions for comprehensive evaluation. Based on the quantitative outcomes, we give a summarization of the merits/demerits of the evaluated algorithms from both performance and efficiency perspectives.
\end{abstract}

\section{Introduction}\label{sec:intro}
\begin{figure}[t]
	\centering
	\includegraphics[width=1.0\linewidth]{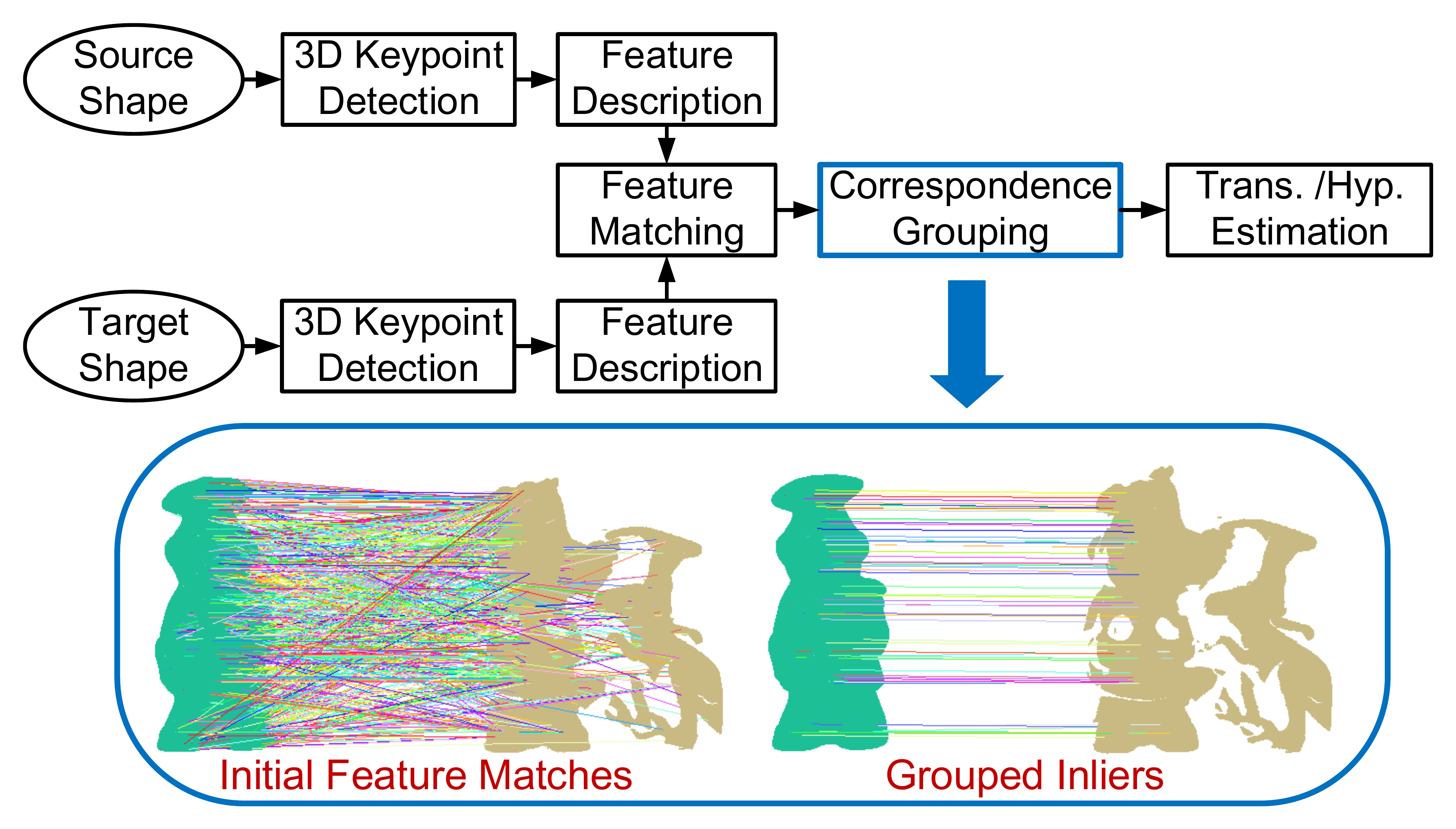}\\
	\caption{Illustration of local feature-based matching paradigm, where the objective of correspondence grouping is searching inliers from the initial matches between two shapes.}
	\label{fig:Def}
\end{figure}

Establishing correct matching relationship between 3D shapes, also known as correspondence problem, is a cornerstone in 3D computer vision. One critical reason is the popularity of local shape feature-based matching paradigm in applications such as 3D object recognition~\cite{guo2013rotational}, point cloud registration~\cite{rusu2009fast}, shape retrieval~\cite{boyer2011shrec} and 3D object categorization~\cite{salti2010use}. Local feature-based matching (Fig.~\ref{fig:Def}) starts from detecting a set of distinctive keypoints on the surface and representing the local shape geometry with feature descriptors, and subsequently generates raw initial matches for recognizing the similarities between two shapes. However, one must expect a high amount of false matches due to two main reasons. One is the residual errors loaded from the former modules, e.g., keypoint localization errors and mismatches of feature descriptors in repetitive structures. The other concerns about nuisances including noise, varying point densities, clutter, occlusion and partial overlaps. To ensure the accuracy of the subsequent transformation estimation or hypothesis generation, \textit{inliers} are desired to be filtered from the raw feature matches, highlighting the importance of correspondence grouping.

A pleasurable correspondence grouping algorithm is amenable to find as many as inliers from the initial feature matches, giving an increase in both precision and recall~\cite{buch2014search}. Similar to the trend in 2D image domain~\cite{cho2009feature,enqvist2009optimal,cho2014finding}, a notable scientific passion has recently characterized the field of 3D correspondence grouping driven by its related higher-level vision tasks such as 3D object recognition and 3D reconstruction.  In addition to the re-exploration of popular 2D correspondence grouping techniques such as similarity score (SS)~\cite{mian2006three,yang2016fast}, nearest neighbor similarity ratio (NNSR)~\cite{lowe2004distinctive}, random sample consensus (RANSAC)~\cite{fischler1981random} and spectral technique (ST)~\cite{leordeanu2005spectral} in 3D domain, we can also find many lately 3D-targeted algorithms such as geometric consistency (GC)~\cite{chen20073d}, clustering~\cite{mian2010repeatability}, game-theory~\cite{rodola2013scale}, 3D Hough voting (3DHV)~\cite{tombari2010object}, and search of inliers (SI)~\cite{buch2014search}. With the wealth of a wide range 3D correspondence grouping algorithms, yet, the effectiveness of these algorithms are usually assessed on datasets of a particular application with limited number of nuisances and comparisons. It is therefore difficult for the developers to choose a proper algorithm given a specific application.

To this end, we present a comprehensive evaluation of seven state-of-the-art 3D correspondence grouping algorithms, i.e., SS, NNSR, RANSAC, ST, GC, 3DHV and SI. This is the first comprehensive evaluation study of 3D correspondence grouping algorithms, to the best of our knowledge, which considers both classical and latest methods with assessment on benchmarks addressing a variety of applications and nuisances. 
The terms \textit{precision} and \textit{recall} are used to measure the quantitative performance, ensuring a balanced examination on the accuracy of the grouped correspondences and the amount of inliers retrieved from the raw feature matches. Also, we take application context into consideration. To be specific, different applications would result to various ratios of inliers and spatial locations of the initial feature matches, mainly due to different categories and degrees of nuisances. To cover these concerns, we deploy our experiments respectively on the Bologna 3D retrieval (B3R)~\cite{tombari2013performance} , UWA 3D object recognition (U3OR)~\cite{mian2006three,mian2010repeatability} and UWA 3D modeling (U3M)~\cite{mian2006novel} datasets to examine these 3D correspondence grouping algorithms. The B3R dataset tests the robustness of the evaluated algorithms with respect to noise and varying point densities, the U3OR dataset concerns clutter and occlusion, and the U3M dataset contains partially overlapped data. All these nuisances have been quantized for a detailed comparison. In a nut shell, the contributions of this paper are mainly twofold:
\begin{itemize}
	\item We give a review and a quantitative evaluation of seven state-of-the-art 3D correspondence grouping algorithms on three benchmarks with various nuisances including noise, point density variation, clutter, occlusion and partial overlaps. The time efficiency regarding different amounts of initial matches is also tested. 
	\item Instructive summarizations including the traits, advantages and limitations of different algorithms are presented.
\end{itemize}

The paper is organized as follows. Sect.~\ref{sec:methods} gives a review of seven state-of-the-art algorithms by identifying the core computational steps of each proposal. Sect.~\ref{sec:eva_meth} shows the evaluation methodology, which consists of the datasets, the performance measures and the implementation details of the evaluated algorithms. The experimental results are reported in Sect.~\ref{sec:exp} , while the conclusions are drew in Sect.~\ref{sec:conc}.

\section{3D Correspondence Grouping Algorithms}\label{sec:methods}
This section briefly reviews several state-of-the-art 3D correspondence grouping algorithms. The correspondence grouping problem can be formulate as: given a source shape  $\cal{S}$  and a target shape $\cal{S'}$, where an initial correspondence set $\cal C$ is generated after  matching the feature sets $\cal F$ and $\cal F'$ respectively extracted on $\cal{S}$ and $\cal{S'}$, the aim is to find a consistent subset of $\cal C$ that identifies the correct matching relationship between $\cal{S}$ and $\cal{S'}$, namely the inlier set ${\cal C}_{inlier}$. A component in $\cal C$ can be parametrized by: $c = \{ p,p', s_{\cal F}(f,f')\} $, with $p \in {\cal S}$, $p' \in {\cal S'}$, $f \in {\cal F}$, $f' \in {\cal F'}$, and $s_{\cal F}(f,f')$ being the feature similarity score assigned to $c$. With these notations, we describe the key ideas and  computation steps of each algorithm in the following.
\\
\\
\noindent\textbf{Similarity Score.} Splitting the initial correspondence set based on the similarity score $s_{\cal F}(f,f')$ is a straightforward solution~\cite{mian2006three,yang2016fast}. It is based on the assumption that correspondences with relatively high precisions possess higher possibility of being correct. Although a number of distinctive 3D local features~\cite{tombari2010unique,guo2013rotational} have been proposed, other disturbances such as noise, missing regions and repetitive patterns could easily cause false judges. This algorithm is served as a baseline in our evaluation, which judges a correspondence as correct if:
\begin{equation}
{1-\| {f - f'} \|_{{L_2}}} \ge {t_{ss}}.
\end{equation}
The popular $L_2$ distance is used to calculate $s_{\cal F}(f,f')$ in this paper.
\\
\\
\noindent\textbf{Nearest Neighbor Similarity Ratio.} Another baseline algorithm evaluated in this paper is Lowe's ratio rule~\cite{lowe2004distinctive}. It penalizes correspondences by the ratio of the nearest and the second-nearest distance in feature space. It enables \textit{distinctive} regions export high ranking scores. Similar to SS's thresholding strategy, NNSR algorithm accepts a correspondence as inlier if:
\begin{equation}\label{eq:ratio}
1 - \frac{{{{\left\| {f - f_1^{'}} \right\|}_{{L_2}}}}}{{{{\left\| {f - f_2^{'}} \right\|}_{{L_2}}}}} \ge {t_{nnsr}},
\end{equation}
with $t_{nnsr} \in [0,1]$.
\\
\\
\noindent\textbf{Random Sample Consensus.} RANSAC~\cite{fischler1981random} is a iterative method which judges the correctness of current samples through the returned number of inliers, and is broadly adopted in both 2D~\cite{brown2007automatic} and 3D domains~\cite{rusu2008aligning}. Despite its variants~\cite{torr2000mlesac,chum2005matching}, we focus on the prototype whose main steps are as follows. 

Given $N_{ransac}$ iterations, at each iteration, the algorithm first randomly samples three components from $\cal C$.  Second, the sampled correspondences are used to compute a transformation ${\bf{T}}_i$. To judge the correctness of ${\bf{T}}_i$, all source keypoints in $\cal C$ (i.e., the points shared by $\cal S$ and $\cal C$) would be transformed using ${\bf{T}}_i$. The confidence of ${\bf{T}}_i$ is positively correlated to  the number of transformed source keypoints whose Euclidean distances to their corresponding points in $\cal S'$ are smaller than a threshold $d_{ransac}$. Finally, the transformation yielding to the maximum inlier count is computed as the optimal ${\bf{T}}^*$, and correspondences in $\cal C$ agreeing with ${\bf{T}}^*$ are grouped as inliers.
\\
\\
\noindent\textbf{Spectral Technique.} Spectral methods are commonly used for searching the main cluster of a graph~\cite{shi2000normalized,mahamud2003segmentation}. Based on the observation that inliers in $\cal C$ should form a consistent cluster, Leordeanu and Hebert~\cite{leordeanu2005spectral} used a spectral technique (ST) to group correspondences. The basic idea is to find the level of \textit{association} of each correspondence with the main cluster exits in the initial correspondence set $\cal C$. In detail, the algorithm operates as follows.

First, a non-negative matrix $\bf M$ comprising all pairwise terms between correspondences in $\cal C$ is built. Second, the principle eigenvector of $\bf M$ is calculated as $\bf v$, and the location of the maximum value of $\bf v$, e.g., ${\bf v}_i$, indicates $c_i$ being inlier. Third, remove from $\cal C$ all potential components in conflict with $c_i$. By repeating step 2 and step3 until ${\bf v}_i =0$ or $\cal C$ is empty, the selected candidates from step 2 thus consist the final inlier set.

ST is generative for both 2D and 3D correspondence problems, depending on how the pairwise term being defined. Here, we use the popular rigidity constrain~\cite{johnson1998surface,buch2014search} in 3D domain as the pairwise term of $c_1$ and $c_2$, which is defined as:
\begin{equation}\label{eq:rigid}
r ({c_1},{c_2}) = \min ( {\frac{{{{\| {{p_1},{p_2}} \|}_{{L_2}}}}}{{{{\| {p'_1,p'_2} \|}_{{L_2}}}}},\frac{{{{\| {p'_1,p'_2} \|}_{{L_2}}}}}{{{{\| {{p_1},{p_2}} \|}_{{L_2}}}}}} ).
\end{equation}
Through thresholding on $r ({c_1},{c_2})$ using $t_{st}$, one can judge whether $c_1$ and $c_2$ are compatible or not.
\\
\\
\noindent\textbf{Geometric Consistency.} The GC algorithm~\cite{chen20073d} is independent from the feature space and applies constrains relating to the compatibility of spatial locations of corresponding points. The compatibility score for two given correspondences $c_1$ and $c_2$ is given as:
\begin{equation}\label{eq:gc}
d({c_1},{c_2}) = | {d({p_1},{p_2}) - d({p'_1},{p'_2})} | < {t_{gc}},
\end{equation}
with $d({p_1},{p_2})={\|p_1 - p_2 \|}_{L_2}$, and $t_{gc}$ being a threshold to judge if $c_1$ and $c_2$ satisfy the geometric constrain or not.

With above rule, the algorithm then associates a \textit{consistent} cluster to each correspondence. Particularly, given a correspondence $c$, its compatibility scores with all other correspondences in $\cal C$ are computed using Eq.~\ref{eq:gc}. All the correspondences with accepted compatibility scores therefore form a cluster for $c$, and the size of the cluster decides the confidence  of current cluster of being the inlier cluster. By repeating the procedure for all correspondences, the biggest cluster  then outputs as the final grouped inlier set.
\\
\\
\noindent\textbf{3D Hough Voting.} The Hough Transform~\cite{vc1962method} is a popular computer vision technique originally proposed to detect lines in images. Tombari and Stefano~\cite{tombari2010object} introduced a 3D extension named 3D Hough voting (3DHV) for object recognition. This method is also employed to group correspondences for partial shape matching~\cite{petrelli2015pairwise}. In 3DHV, each correspondence casts a vote in 3D Hough space based on the following steps.

For the $i$th correspondence in $\cal C$ denoted by $c_i = \{ p_i,p'_i\} $, the vector between $p_i \in {\mathbb{R}^3} $ and the  centroid $C_{\cal S} \in {\mathbb{R}^3}$ of the source shape $\cal S$ is firstly computed as:
\begin{equation}
{\bf{V}}_{i,G}^{\cal S} = {C_{\cal S}} - {p_i},
\end{equation}
which is then transformed in the coordinates given by the local reference frame (LRF) of $p_i$ as:
\begin{equation}
{\bf{V}}_{i,L}^{\cal S} = {\bf{R}}_i^{\cal S} \cdot {\bf{V}}_{i,G}^{\cal S},
\end{equation}
where ${\bf{R}}_i^{\cal S}$ is the rotation matrix and each line in ${\bf{R}}_i^{\cal S}$ is a unit vector of the LRF of $p_i$. Note that LRF is an independent coordinate system established in the local surface around the keypoint, and many 3D feature descriptors~\cite{tombari2010unique,guo2013rotational} provide LRF for feature representation. This step endows the vector of $p_i$ with invariance to rigid transformation. Analogously, we can obtain a vector ${\bf{V}}_{i,L}^{\cal S'}$ for $p'_i$. If $p_i$ and $p'_i$ are correctly corresponded, ${\bf{V}}_{i,L}^{\cal S'}$ should coincide with ${\bf{V}}_{i,L}^{\cal S}$. Based on this assumption, the vector ${\bf{V}}_{i,L}^{\cal S'}$ is finally transformed in the global coordinate of $\cal S'$ as:
\begin{equation}
{\bf{V}}_{i,G}^{\cal S'} = {\bf{R}}_i^{\cal S'} \cdot {\bf{V}}_{i,L}^{\cal S'}+p'_i.
\end{equation}
With these transformations, the feature $f'_i$ could vote in a 3D Hough space by means of a vector ${\bf{V}}_{i,G}^{\cal S'}$. The peak in the Hough space indicates the cluster constituted by inliers. 
\\
\\
\noindent\textbf{Search of Inliers.} The search of inliers (SI)~\cite{buch2014search} algorithm is a recent proposal targeting at solving 3D correspondence problem. The core idea is a combination of local and global constrains to determine if a vote should be cast. We summarize this algorithm into three main steps, i.e, initialization, local voting and global voting.

During initialization, a subset of the initial correspondence set $\cal C$ is extracted using the Lowe's ratio rule (c.f. Eq.~\ref{eq:ratio}) as ${\cal C}_{Ratio}$. At the local voting stage, the shared correspondences of ${\cal C}_{Ratio}$ and the nearest $\kappa$ correspondence neighbors of $c$ are defined as the local voters for $c$, denoted by ${\cal C}_L(c)$. The components in ${\cal C}_L(c)$ that satisfy the rigidity constrain (c.f. Eq.~\ref{eq:rigid}) are defined as the positive local votes $\Upsilon_L(c)$:
\begin{equation}
\Upsilon_L(c)=\{c_L \in {\cal C}_L(c) : r(c,c_L)>\varsigma\},
\end{equation}
where $\varsigma$ is a free parameter in local voting stage. The local score of $c$ is then defined as $s_L(c)=\frac{|\Upsilon_L(c)|}{|{\cal C}_L(c)|}$.

At the global voting stage, the global voters ${\cal C}_G$ are selected as the former $\kappa$ correspondences ranked using the Lowe's ratio score in a decreasing order. To judge the affinity between two correspondences $c_1$ and $c_2$, the following test is prepared:
\begin{equation}
v_G(c_1,c_2)=d({\bf T}(c_1) \cdot p_2, p'_2),
\end{equation}
with ${\bf T}(c) $ defined as ${\bf{R}(p')}^{-1} \cdot {\bf{R}(p)}$, where ${\bf{R}(p)}$ represents the LRF of $p$. The global votes are then found by applying both local and global constrain as:
\begin{equation}
\Upsilon_G(c)=\{c_G \in {\cal C}_G : r(c,c_G)>\varsigma  \wedge v_G(c,c_G) < \delta \},
\end{equation}
where $\delta$ is a Euclidean distance tolerance. The eventual vote score for $c$ is defined as:
\begin{equation}
s(c)=\frac{|\Upsilon_L(c)|+|\Upsilon_G(c)|}{|{\cal C}_L(c)|+|{\cal C}_G(c)|}.
\end{equation}

By thresholding on $s(c)$ based on Otsu's adaptive method~\cite{otsu1975threshold}, the correspondences left remain SI-judged inliers.

\section{Evaluation Methodology}\label{sec:eva_meth}
All the algorithms presented in Sect.~\ref{sec:methods} have been evaluated on three chosen benchmarks, where different levels of noise, point density variation, clutter, occlusion and partial overlap are contained. The calculated inliers by all algorithms are measured using the \textit{precision} and \textit{recall} criteria. This section also presents the implementation details of each algorithm.
\subsection{Datasets}
\begin{figure}[t]
	\centering
	\includegraphics[width=1.0\linewidth]{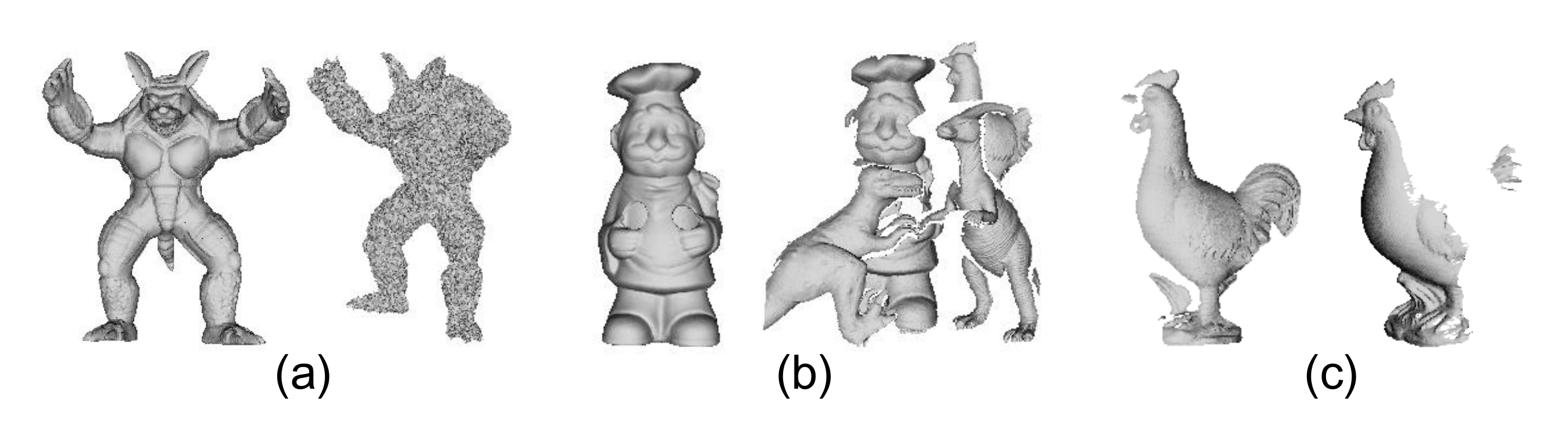}\\
	\caption{Sample views (visualized in mesh) from (a) B3R, (b) U3OR and (c) U3M datasets.}
	\label{fig:dataset}
\end{figure}
\noindent\textbf{B3R Dataset.} The Bologna 3D Retrieval (B3R) dataset~\cite{tombari2013performance}, with 8 models and 18 scenes, is considered to test the algorithms' robustness with respect to various levels of noise and varying point densities. The models are taken from the Stanford Repository\footnote{www.graphics.stanford.edu/data/3Dscanrep}, while the scenes are the rotated copies of these models. In specific, we inject the scenes with Gaussian noise along the $x$, $y$ and $z$ axes. The standard deviation of noised increases from 0.05 to 0.45\textit{pr} with in incremental step of 0.05\textit{pr}, where \textit{pr} hereinafter denotes the point cloud resolution, i.e., the average shortest distance among neighboring points in the point cloud. Further, we down-sample the scenes from 0.9 to 0.1 data resolution with an interval of 0.1 data resolution. The enriched B3R dataset then incorporates 324 scenes with quantized levels of noise and point density variation.
\\
\\
\noindent\textbf{U3OR Dataset.} The UWA 3D object recognition (U3OR) dataset~\cite{mian2006three,mian2010repeatability} is a popular benchmark for 3D object recognition~\cite{guo2013rotational,guo2016comprehensive}, where 5 models and 50 scenes are included. Each scene contains four or five models in the presence of approximately 65\%-95\% degrees of clutter and 60\%-90\% occlusion. A total of 188 valid matching instance pairs can be found in this dataset, whose objective is to test the correspondence grouping algorithms' resilience to clutter and occlusion.
\\
\\
\noindent\textbf{U3M Dataset.} The UWA 3D modeling (U3M)~\cite{mian2006novel} dataset belongs to the point cloud (2.5D view) registration scenario. There are 22, 16, 16 and 21 2.5D views respectively captured from the \textit{Chef}, \textit{Chicken}, \textit{T-rex}, and \textit{Parasaurolophus} models. We obtain the ground truth transformations of each considered data pair via first manually alignment and then iterative closest points (ICP)~\cite{besl1992method} refinement. We finally screen out 340 valid pairs from this dataset with 30\%-90\% degrees of overlap.
\subsection{Criteria}
Let ${\bf{T}}_{GT} =\{ {\bf{R}}_{GT},  {\bf{t}}_{GT}  \}$ denote the ground truth transformation between $\cal S$ and $\cal S'$, where ${\bf{T}}_{GT}\in SE(3)$, ${\bf{R}}_{GT}\in SO(3)$ and ${\bf{t}}_{GT}\in {\mathbb{R}^3}$. A correspondence $c=(p,p')$ is accepted as correct only if:
\begin{equation}\label{eq:judge}
{\|  p\cdot {\bf{R}}_{GT} + {\bf{t}}_{GT} - p'  \|}_{L_2} \le \epsilon,
\end{equation}
where $\epsilon$ is a judging threshold. Let ${\cal C}_{inlier}$, ${\cal C}_{inlier}^{corret}$ and ${\cal C}_{inlier}^{GT}$ respectively represent the grouped inlier set, the correct judged inliers in the grouped set, and the ground truth inlier set in the initial correspondence set $\cal C$, we measure the quality of an algorithm using \textit{precision} and \textit{recall} defined as:
\begin{equation}
{\rm{Precision}} =\frac{|{\cal C}_{inlier}^{corret}|}{|{\cal C}_{inlier}|},
\end{equation}
\begin{equation}
{\rm{Recall}} =\frac{|{\cal C}_{inlier}^{corret}|}{|{\cal C}_{inlier}^{GT}|},
\end{equation}
where $|\cdot|$ denotes the cardinality of a set.
\subsection{Implementation Details}
\begin{table}[t]\small
	\renewcommand{\arraystretch}{1}
	\caption{Parameters used through the evaluation.}
	\label{tab:para}
	\centering
	\begin{tabular}{@{\extracolsep{\fill}}lll}
		\hline
		SS  & $t_{ss}$ & Adaptive~\cite{otsu1975threshold} \\
		\hline
		NNSR & $t_{nnsr}$ & 0.8 \\
		\hline
		RANSAC & $N_{ransac}$ & 10000 \\
		& $d_{ransac}$ & 5\textit{pr}\\
		\hline
		ST & $t_{st}$ & 0.6\\
		\hline
		GC & $t_{gc}$ & 3\textit{pr}\\ 
		\hline
		3DHV & - &- \\
		\hline
		SI & $\kappa$ & 250 \\
		& $\varsigma$ & 0.9 \\
		& $\delta$ & 5\textit{pr}\\
		\hline
	\end{tabular} 
\end{table}
The input for the algorithms evaluated in this paper, i.e., the initial correspondence set $\cal C$, is generated via Harris 3D~\cite{sipiran2011harris} keypoint detection, SHOT~\cite{tombari2010unique} feature description and $L_2$ distance-based feature matching~\cite{guo2014accurate,yang2016fast}. In default setting, we set the Non-Maxima-Suppression radius of Harris 3D detector as 3\textit{pr}, generating around 1000 keypoints for a point cloud containing a hundred thousand of points. The support radius of SHOT is 15\textit{pr} as suggested in~\cite{guo2013rotational}, while the judging threshold $\epsilon$ equals to 4\textit{pr}.

Regarding the parameters of each algorithm, we list them in Table~\ref{tab:para}. Notably, we make $t_{ss}$ adaptive using~\cite{otsu1975threshold} because a fixed value can be hardly turned towards feature matches with different qualities. The thresholds in NNSR and SI algorithms are kept consistent to those in their original papers. For ST and GC algorithms, their thresholds are determined via tuning experiments. In RANSAC, considering the magnitude of the initial correspondence set, 10000 loops are assigned to strike a balance between effectiveness and efficiency.

All the algorithms are implemented in C++ with the help of point cloud library (PCL)~\cite{rusu20113d}, using a PC equipped with a 3.4GHz processor and 24GiB memory.
\section{Experimental Results}\label{sec:exp}
\begin{figure*}[t]
	\begin{minipage}{0.5\linewidth}
		\centering
		\subfigure[Noise]{
			\includegraphics[width=1.0\linewidth]{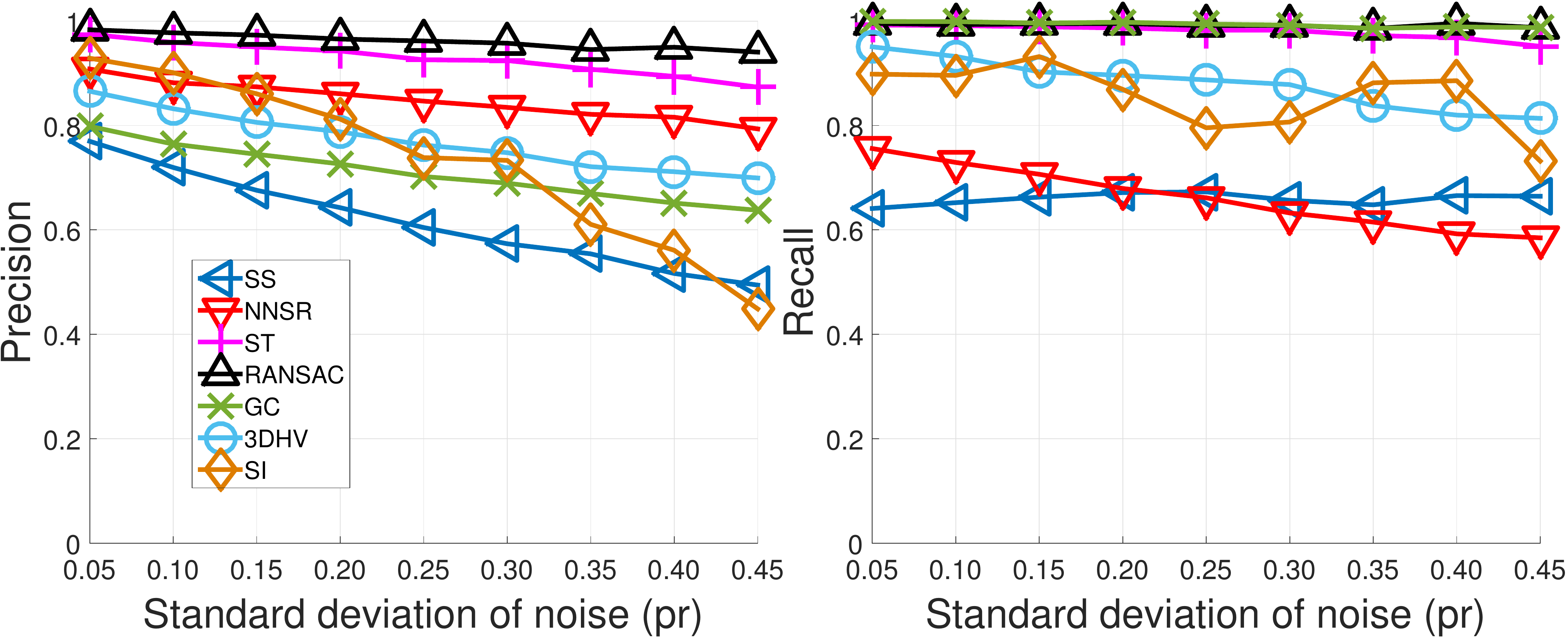}}
	\end{minipage}
	\hfill
	\begin{minipage}{0.5\linewidth}
		\centering
		\subfigure[Point density variation]{
			\includegraphics[width=1.0\linewidth]{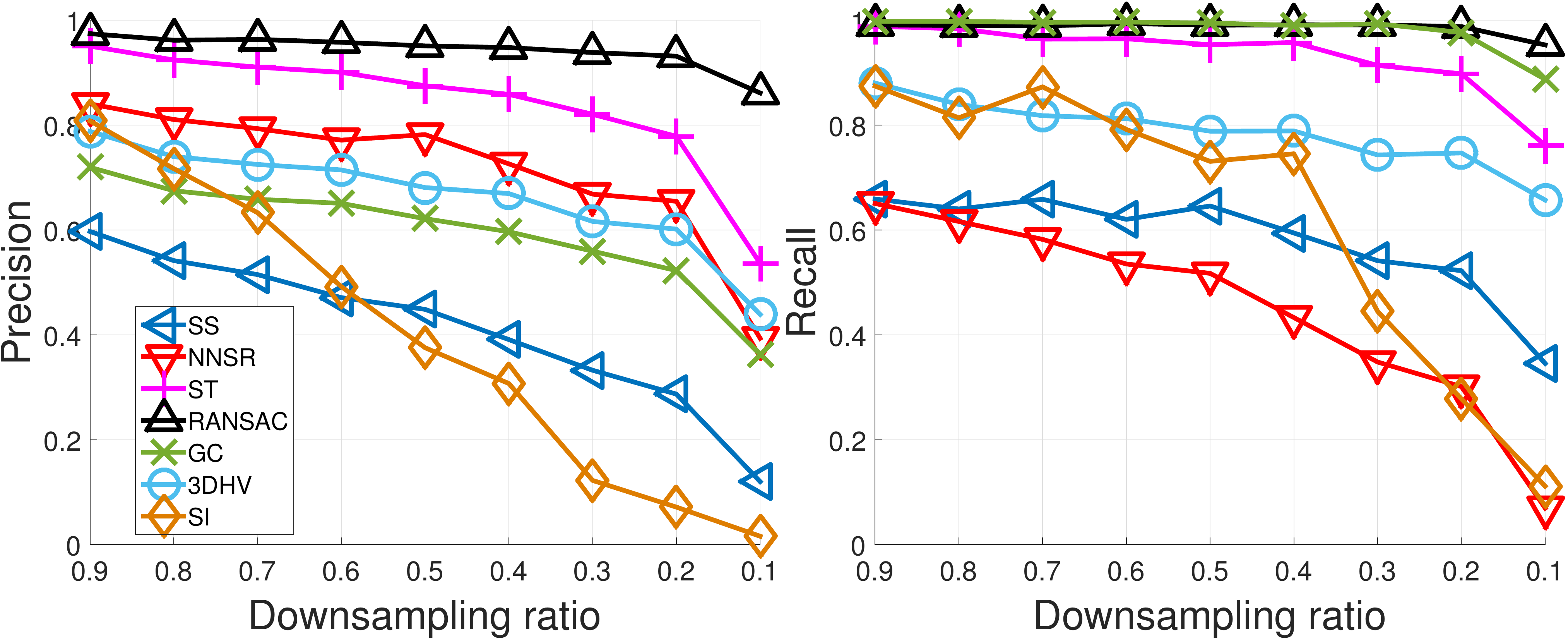}}
	\end{minipage}
	\hfill
	\begin{minipage}{0.5\linewidth}
		\centering
		\subfigure[Clutter]{
			\includegraphics[width=1.0\linewidth]{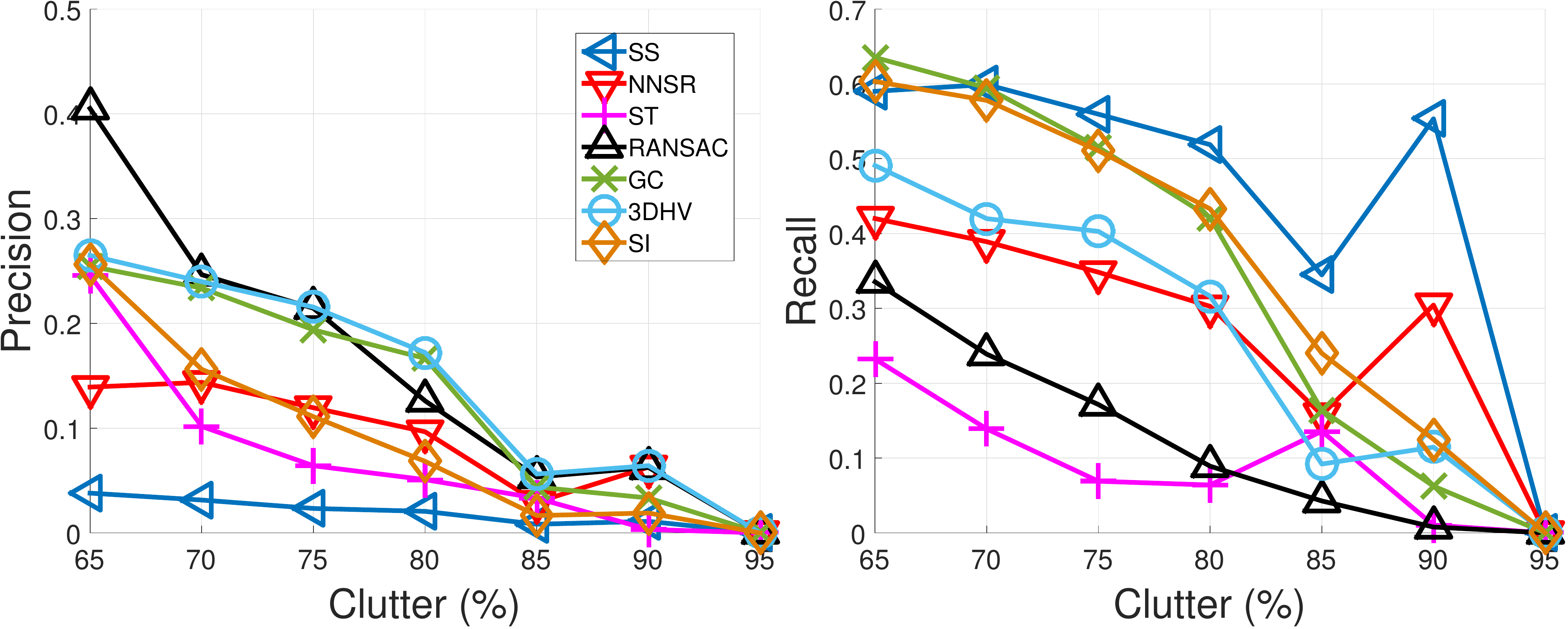}}
	\end{minipage}
	\hfill
	\begin{minipage}{0.5\linewidth}
		\centering
		\subfigure[Occlusion]{
			\includegraphics[width=1.0\linewidth]{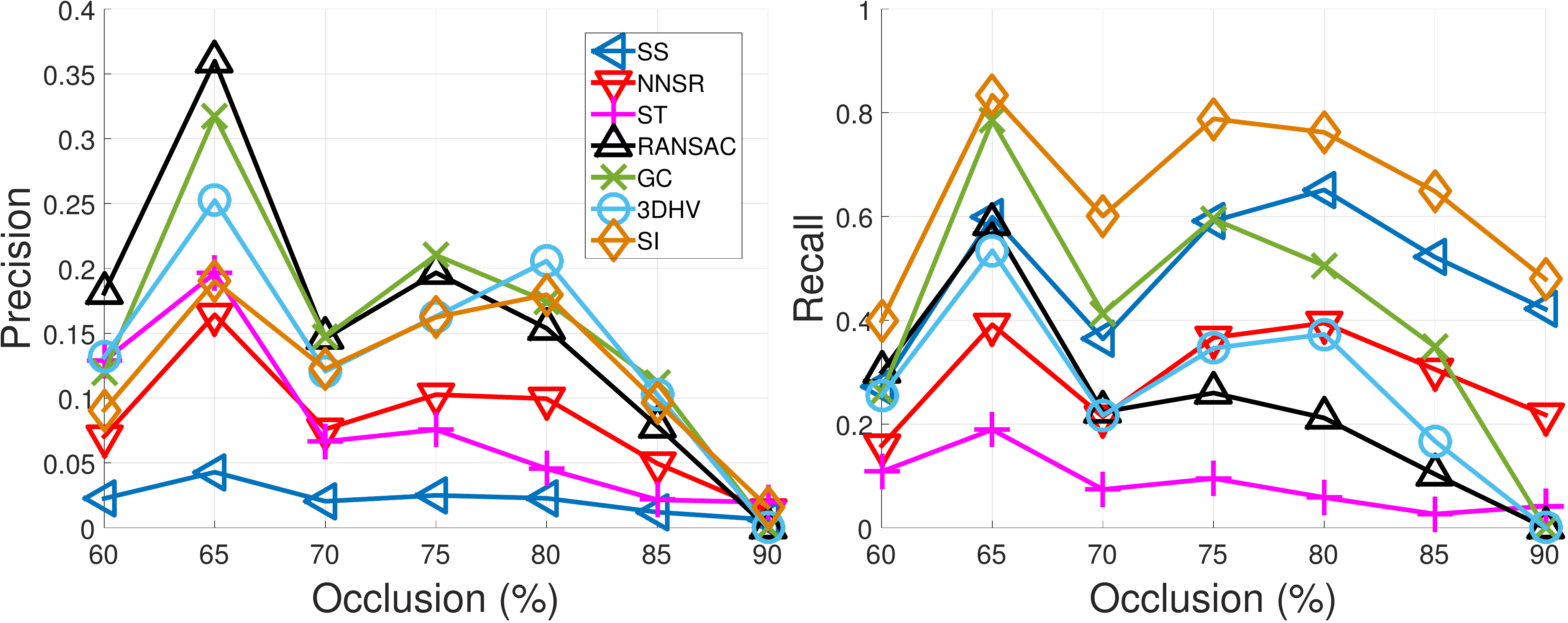}}
	\end{minipage}
	\hfill
	\begin{minipage}{0.5\linewidth}
		\centering
		\subfigure[Partial overlap]{
			\includegraphics[width=1.0\linewidth]{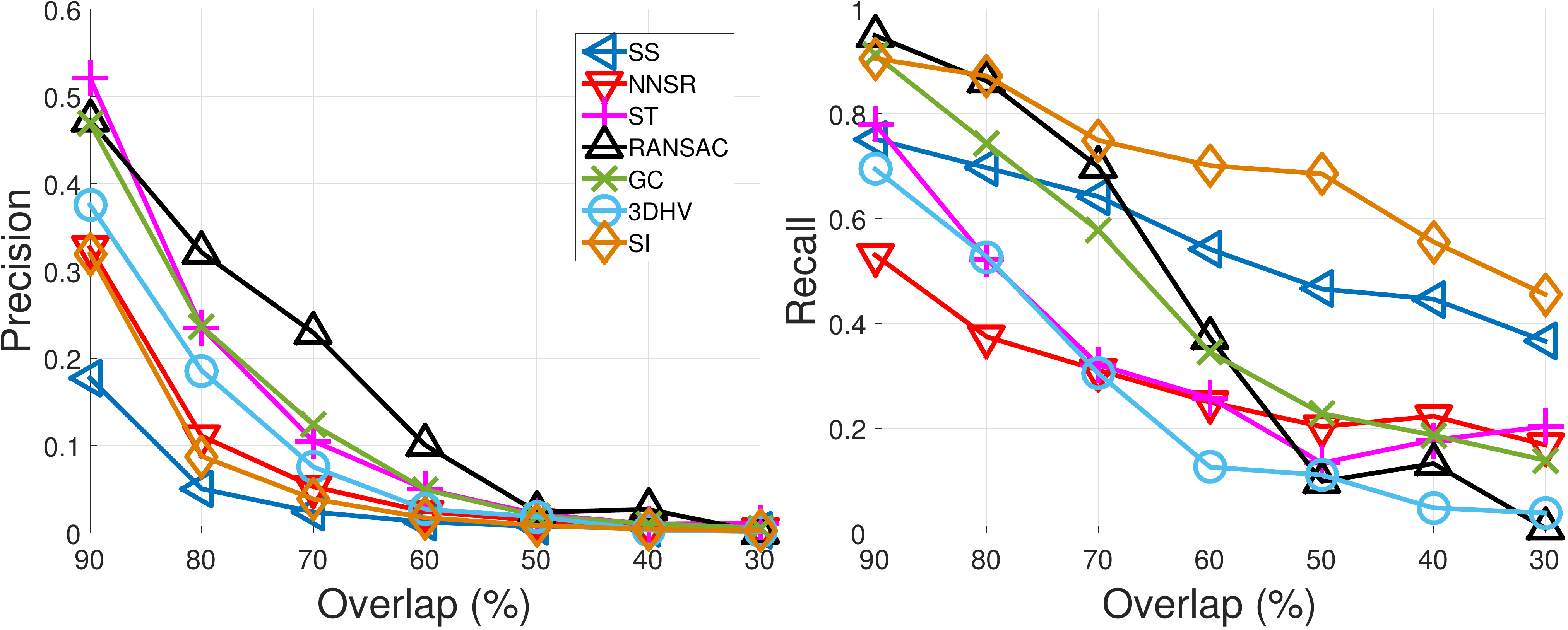}}
	\end{minipage}
	\hfill
	\begin{minipage}{0.5\linewidth}
		\centering
		\subfigure[Threshold $\epsilon$]{
			\includegraphics[width=1.0\linewidth]{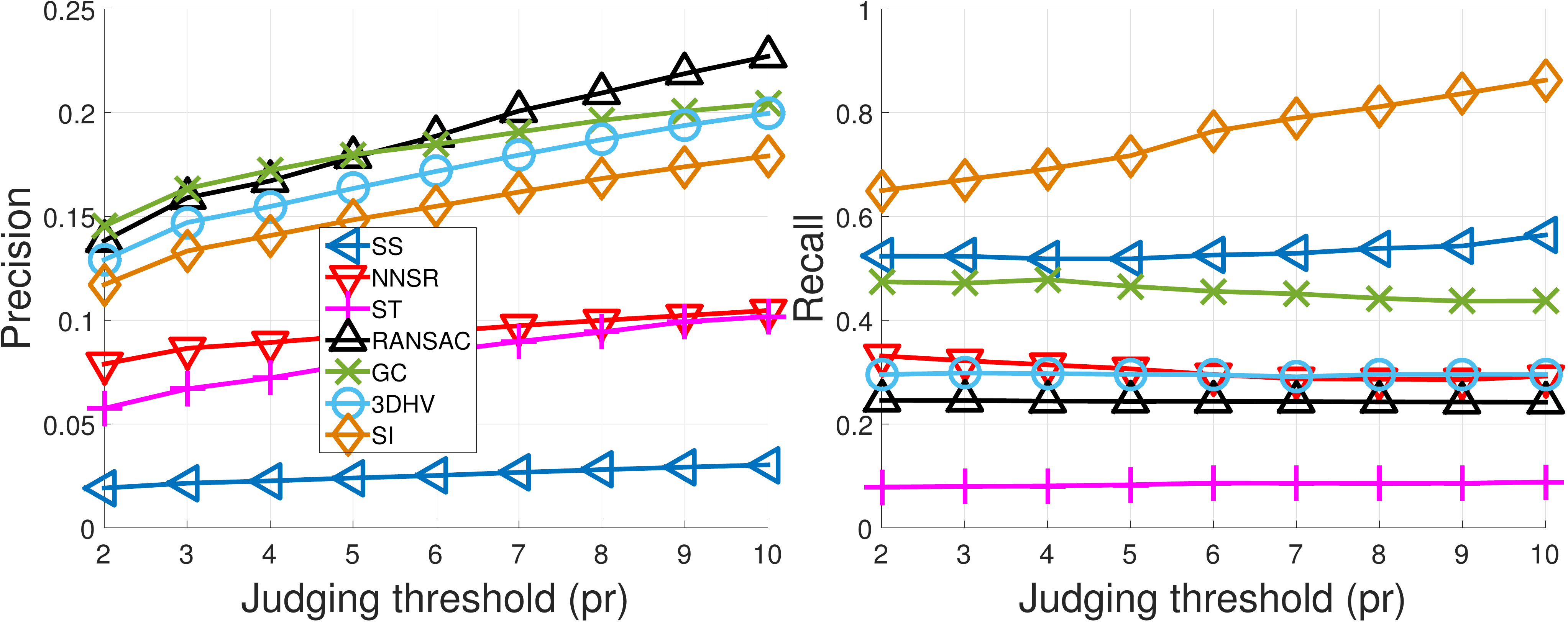}}
	\end{minipage}
	\hfill
	\begin{minipage}{0.5\linewidth}
		\centering
		\subfigure[Number of initial correspondences]{
			\includegraphics[width=1.0\linewidth]{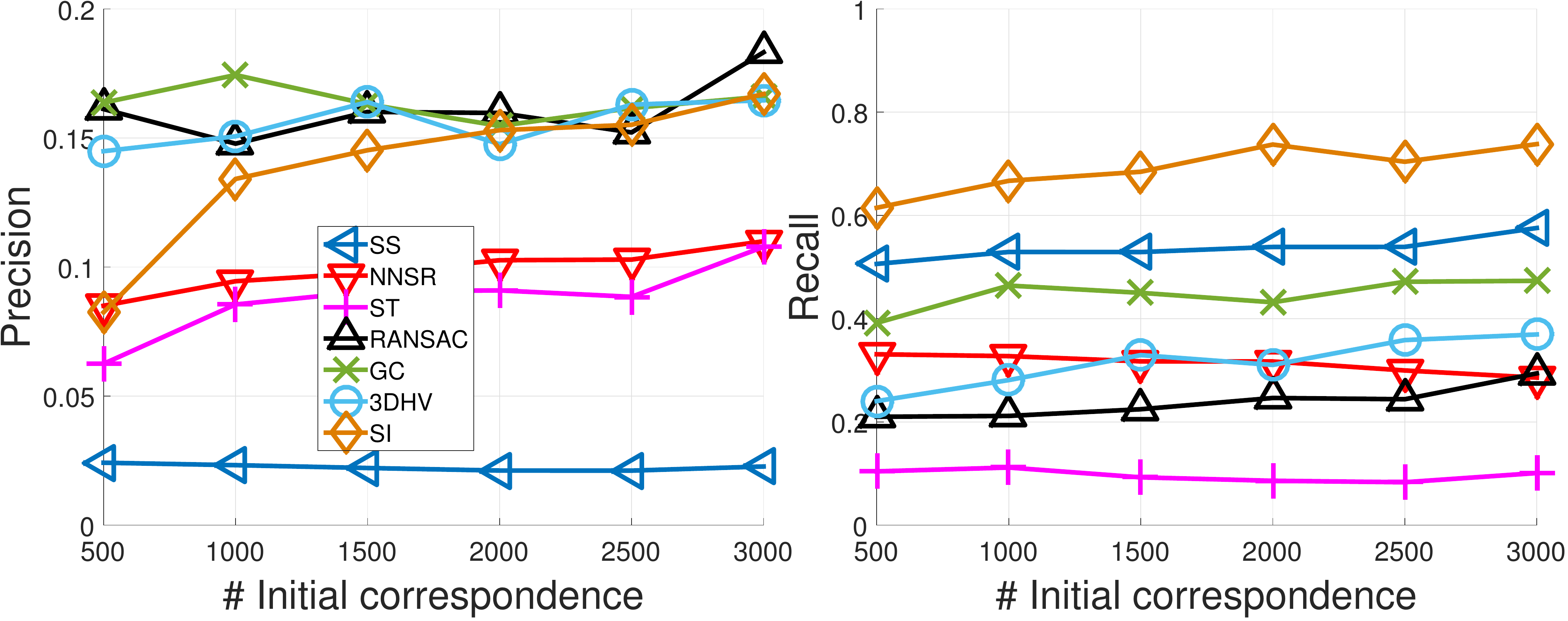}}
	\end{minipage}
	\hfill
	\begin{minipage}{0.42\linewidth}
		\centering
		\subfigure[Time efficiency]{
			\includegraphics[width=1.0\linewidth]{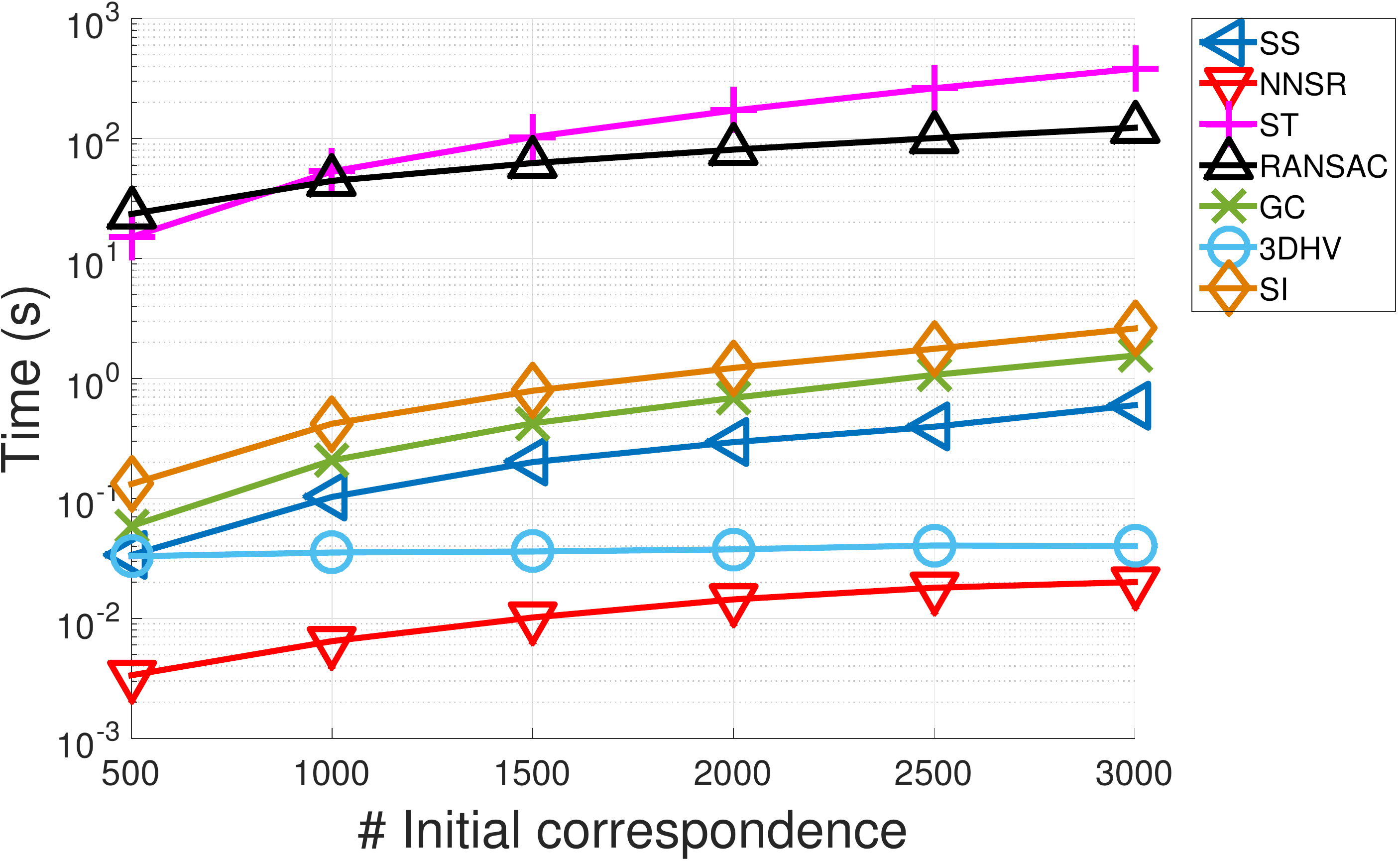}}
	\end{minipage}
	\hfill
	\caption{(a-g): \textit{Precision} and \textit{Recall} performance of seven correspondence grouping algorithms with respect to different nuisances on the experimental datasets. (h): Time efficiency performance regarding different sizes of the initial correspondence set, where the $y$-axis is logarithmic for best view.}
	\label{fig:performance}
\end{figure*}
This section provides the outcomes of each evaluated algorithm (Sect.~\ref{sec:methods}) on the experimental datasets using the protocols in Sect.~\ref{sec:eva_meth}. The assessed terms including: robustness to noise, varying point densities, clutter, occlusion, partial overlap, varying threshold $\epsilon$ (c.f. Eq.~\ref{eq:judge}), varying sizes of the initial correspondence set, and computational cost.
\subsection{Performance on the B3R Dataset}
\noindent\textbf{Robustness to noise.} Noise is expected to have an impact on the discriminative power of the feature descriptor, thus creating a certain amount of false matches. While in retrieval context, the ratio of inliers is generally high because the models used in the B3R dataset are of rich geometric information. The result in such context is shown in Fig.~\ref{fig:performance}(a).

As witnessed by the figure, RANSAC and SI appear to be the best two ones among all evaluated proposals, considering their overall precision and recall performance. An interesting finding is that NNSR even surpasses GC and 3DHV in cases with extreme noise in terms of precision. That is because NNSR prefers to select \textit{distinctive} correspondences, which is peculiar sufficient in this dataset as the models possess wealthy distinctive structures. While the recall of NNSR stays lower than other algorithms except SS. The SI algorithm, with pleasurable performance when the standard deviation of Gaussian noise is less than 0.15\textit{pr}, meets a significant deterioration when the noise turns further severe, indicating its sensitivity to high levels of Gaussian noise.
\\
\\
\noindent\textbf{Robustness to point density variation.}  Similar to noise, this term also affects a descriptor's distinctiveness. We present the result under varying point densities in Fig.~\ref{fig:performance}(b).

We can observe that the behaviors of these algorithms under the effect of point density variation are analogous to those under the challenge of noise. For instance, RANSAC and ST again give the best overall performance, followed by NNSR, 3DHV and GC. Yet, the difference behind is that SS even outperforms SI when the downsampling ratio reaches 0.3 regarding precision. While the recall performance of both NNSR and SI drops dramatically in low-resolution case. That is because SHOT is sensitive to varying point densities~\cite{tombari2010unique}, making the feature be weakly distinctive (e.g., NNSR's principle) for data with high ratios of resolution decimation. While the reason for SI is that SHOT's LRF (e.g., the component in the global voting stage for SI) is less repeatable when faced with data resolution variation~\cite{guo2013rotational}.
\subsection{Performance on the U3OR Dataset}
\noindent\textbf{Robustness to clutter.} The degree of clutter, as defined in~\cite{mian2006three}, is the percentage of non-model surface patch area in the scene. Surface patches in the clutter area with similar geometric properties to the patches in the model would cause outliers during feature matching. The result with quantized levels of clutter is shown in Fig.~\ref{fig:performance}(c).

A clear degrade of performance can be found for all algorithms, as 3D object recognition scenario is  more challenging than retrieval~\cite{guo2016comprehensive}. When the degree of clutter is less than 75\%, RANSAC achieves the best precision performance. As the degree of clutter further increases, 3DHV gives the best performance. Notably, the ST algorithm, with top-ranked performance on the B3R dataset, performs quite poor on the U3OR dataset. That is because ST tries to find large isometry-maintained clusters, which rarely exit in scenes with high percentages of clutter. In terms of recall performance, SS, SI and GC perform better than others. Weighing up both precision and recall, 3DHV and GC are two most superior algorithms under the effect of clutter.
\\
\\
\noindent\textbf{Robustness to occlusion.} Occlusion would result to incomplete shape patches, imposing great challenges for accurate feature description. The degree of occlusion is given as the ratio of occluded model surface patch to the total model surface area~\cite{mian2006three,mian2010repeatability}.

As shown in Fig.~\ref{fig:performance}(d), when the degree of occlusion is smaller than 70\%, RANSAC is the best one regarding precision performance. As the occlusion degree increases to 75\%, GC outperforms RANSAC to be the best. GC, SI and 3DHV eventually surpass other algorithms when the occlusion degree exceeds 75\%.  As for the recall performance, SI outperforms all other algorithms for all levels of occlusion, especially in highly occluded scenes. SS and ST remain to be two poorly performed algorithms in this test. We can infer that consistency-based algorithms, such as RANSAC and GC, are more suitable for scenes with occlusion. While algorithms relying on initial feature matching score, e.g., SS and SI, are of high-risk in dealing with false matches, as the feature matching score measured from occluded scene patches can be suspicious.
\subsection{Performance on the U3M Dataset}
\noindent\textbf{Robustness to partial overlap.} The U3M dataset provides matching pairs with various degrees of overlap. The degree of overlap is measured as the ratio of the number of corresponding vertices to the minimum number of vertices of two shapes~\cite{mian2006novel}.  The result with respect to different overlapping degrees is presented in Fig.~\ref{fig:performance}(e).

Common to all algorithms is that their performance generally degrades as the degree of overlap drops. This is owing to the fact that the ratio of outliers in the initial correspondence set is closely correlated to the ratio of overlapping regions.  As for precision performance, RANSAC generally exceeds the others for all levels of overlapping degrees by a large margin in the range of 60\% to 80\% overlapping degree. GC and ST behave comparable, followed by 3DHV, NNSR, SI and SS. Regarding recall performance, SI is superior to others especially when the degree of overlap is smaller than 70\%.
\subsection{Performance w.r.t. Varying Threshold $\epsilon$}
\begin{figure}[t]
	\centering
	\includegraphics[width=1.0\linewidth]{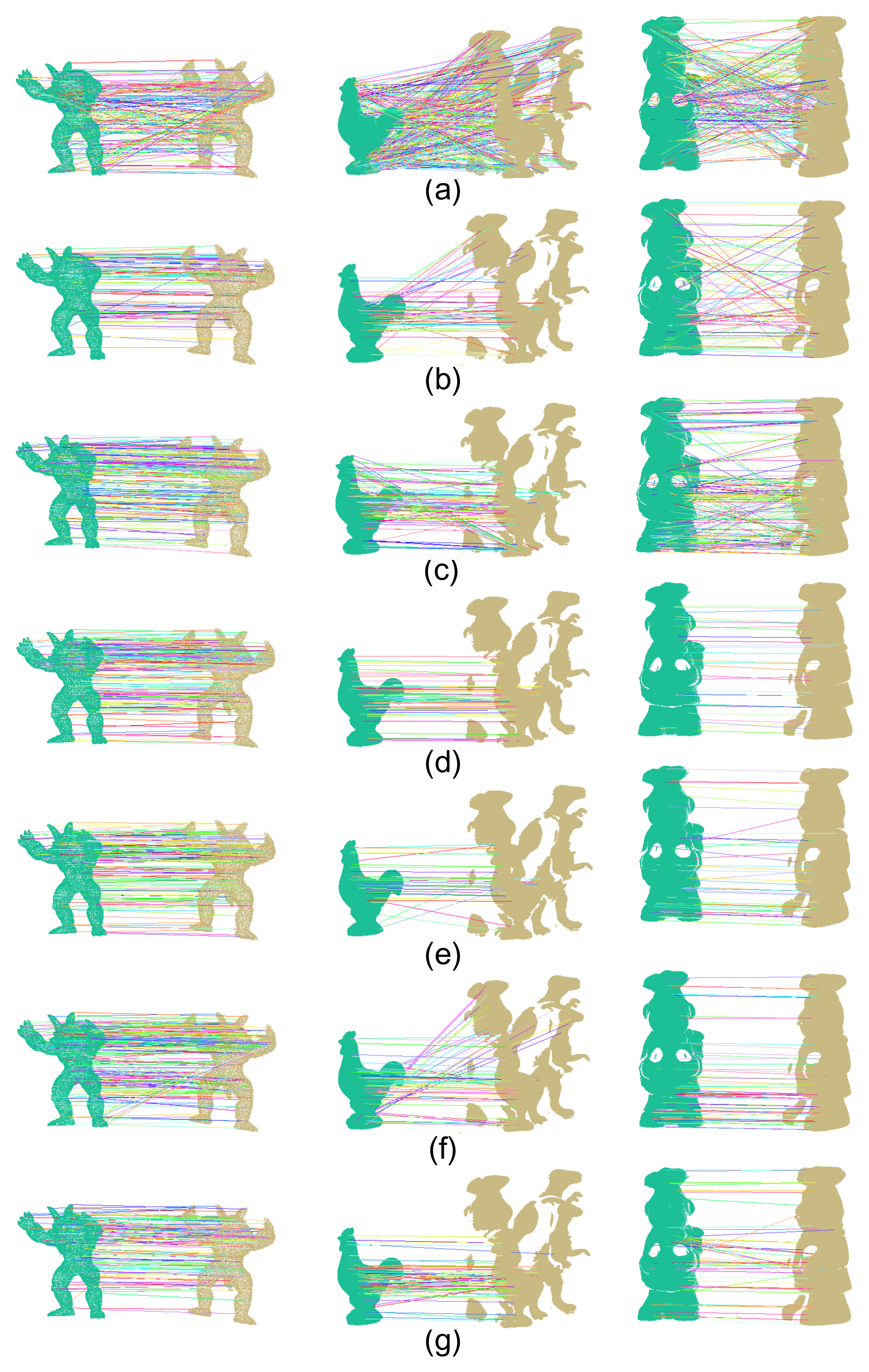}\\
	\caption{Exemplar visual results of the evaluated correspondence grouping algorithms, i.e., (a) SS, (b) NNSR, (c) ST, (d) RANSAC, (e) GC, (f) 3DHV and (g) SI,  respectively from the B3R, U3OR and U3M datasets (from left to right).}
	\label{fig:visual_result}
\end{figure}
As afore defined in Eq.~\ref{eq:judge}, the threshold $\epsilon$ determines that to what extend we judge a correspondence as an inlier. We hereby change this threshold (the default setting is 4\textit{pr}) to examine the performance variation of the evaluated algorithms. Specifically, we conduct this experiment on the whole U3OR dataset, with outcomes being presented in Fig.~\ref{fig:performance}(f).

As expected, all algorithms attain higher precision results with looser $\epsilon$. In particular, GC and RANSAC respectively reach the highest precision when the threshold $\epsilon$ is in the range of [2\textit{pr},5\textit{pr}] and [6\textit{pr},10\textit{pr}]. The precision of SS improves faintly, indicating the majority of its judged inliers deviates a lot from the ground truth inliers. In terms of the recalled inliers, SI and SS present an increasing trend as the threshold gets large, while the performance of other algorithms remains almost unchanged.
\subsection{Performance w.r.t. Varying Numbers of Initial Correspondences}
Various numbers of initial correspondences are desired respecting different applications, such as \textit{dense} matching for shape morphing~\cite{alexa2002recent} and \textit{sparse} matching for crude scan alignment~\cite{rusu2008aligning}. Towards this end, we test the performance of these algorithms with respect to different numbers of initial correspondences on the U3OR dataset, as shown in Fig.~\ref{fig:performance}(g).

The figure suggests, that different algorithms give different responses when varying the number of initial feature matches. The performance of some algorithms, e.g., GC, RANSAC and 3DHV, fluctuates as the number of initial matches augments. Meanwhile, one can find that the size of initial correspondence set has a relatively strong impact on the SI and ST algorithms. To be more specific, when the number of initial correspondences is smaller than 1000, these two algorithms produce low precision. However, as the initial feature matches become dense, i.e., more than 1000 correspondences, the precision performance of SI and ST climbs quickly. Note that the SI algorithm even reaches the second best precision with about 3000 initial correspondences. This is owing to the reason that dense initial correspondences could provide more reliable components in the local consolidation voting set for SI. Still, SI achieves the best recall performance under all tested sizes of the initial correspondence set, surpassing all others by a large gap.
\subsection{Time Efficiency}
In addition to above \textit{precision} and \textit{recall} results, we test the computational efficiency of each evaluated algorithm with respect to different sizes of the input feature matches. The deployment of this experiment is as follows. First, the NMS radius of the Harris 3D keypoint detector is varied to obtain different quantities of initial feature matches. Second, these initial correspondence sets are fed to the evaluated algorithms and their computational costs are recorded. Finally, we repeat the former stage 10 times to eliminate randomness and the average timing results are collected, as shown in Fig.~\ref{fig:performance}(h).

One can make several observations from the results. First, ST and RANSAC are two most time-consuming ones, especially for large initial correspondence sets. The reason for ST is that the computational cost for solving the principle eigenvector of an $n \times n$ matrix increases dramatically when the order $n$ (i.e., the size of the initial correspondence set) gets larger. The explanation for RANSAC is that RANSAC requires a huge amount of iterations to promise acceptable result, while each iteration takes the whole correspondence set into consideration for computing current inliers. Second, NNSR and 3DHV are two most efficient ones. We remark that as SS employs an adaptive thresholding strategy~\cite{otsu1975threshold} in our implementation, its time cost therefore turns to be more expensive than NNSR. The core operation manner of 3DHV is coordinate transformation, it therefore requires very few run time even for thousands of initial correspondences. Third, GC and SI are middle-ranked ones among all evaluated algorithms in terms of run time. The main timing cost of GC is dedicated to compute the distance constrains of all correspondence pairs, while SI needs both local and global consolidations to judge the correctness of a correspondence.
\subsection{Visual Results}
We finally provide some visual results of the evaluated correspondence grouping algorithms in Fig.~\ref{fig:visual_result}. From the figure, we can percept some visual differences of these outcomes. For instance, the number of outliers in the results of the two baseline algorithms, i.e., SS and NNSR, is relatively large except on the B3R dataset. This verifies that algorithms relying on feature matching score are very sensitive to nuisance directly affecting a feature's discriminative ability, e.g., clutter, occlusion, holes and etc.  Another observation is that with different grouping principles, the number as well as the spatial locations of the results of these algorithms generally differ from each other.
\section{Conclusions}\label{sec:conc}
This paper has presented a thorough evaluation of 3D correspondence grouping algorithms on a variety of datasets. The evaluated terms including the precision and recall performance under various levels of noise, point density variation, clutter, occlusion, partial overlap, inlier judging threshold, the size of initial feature matches, and computational efficiency. In light of these evaluation outcomes, we summarize the findings of this paper into several points as follows.
\begin{itemize}
	
	\item SS and NNSR, as two baselines relying on feature matching similarity, is very sensitive to disturbances including clutter, occlusion and partial overlap. Given high quality shapes with rich geometric structures, NNSR can be an effective option which also affords real-time performance.

	\item The ST algorithm is effective for correspondence set with very high amount of inliers, while its performance degrades dramatically under challenging circumstances, e.g., 3D object recognition and 2.5D view matching. Also, ST is shown to be very time-consuming, especially for large-scale correspondence problems.

	\item RANSAC shows superior precision performance under a variety of nuisances, at the expense of relatively long execution time. Hence, RANSAC is suitable for off-line applications relying sparse matching such as scan registration and 3D modeling.

	\item 3DHV is an ultra efficient algorithm which simultaneously returns acceptable inlier searching performance in many applications.  These merits suggest that 3DHV can be applied to time-crucial applications, e.g., simultaneous localization and mapping (SLAM), object grasping and 3D object recognition in robotics.

	\item For applications requiring dense feature correspondences, SI would be the best choice. A core shortcoming of SI is its limited precision under the nuisances of clutter and partial overlap. GC, in this context, can be an alternative which shows overall higher precision.
\end{itemize}

It is noteworthy that although existing algorithms work well under retrieval context even with severe noise and data resolution decimation, their performance is quite limited under clutter, occlusion and partial overlap. We believe the research should towards the development of robust correspondence grouping algorithms for 3D object recognition and point cloud registration applications. 
\\\\
\noindent\textbf{Acknowledgment.}
The authors would like to acknowledge the Standford 3D Scanning Repository, the University of Western Australia and the University of Bologna for providing their datasets. We also thank Dr. Buch for sharing the code to us. This work is jointly supported by the National High Technology Research and Development Program of China (863 Program) under Grant 2015AA015904 and the 2015 annual foundation of China Academy of Space Technology (CAST).

{\small
	\bibliographystyle{ieee}
	\bibliography{mybibfile}

\begin{thebibliography}{10}\itemsep=-1pt

\bibitem{alexa2002recent}
M.~Alexa.
\newblock Recent advances in mesh morphing.
\newblock In {\em Computer Graphics Forum}, volume~21, pages 173--198. Wiley
  Online Library, 2002.

\bibitem{besl1992method}
P.~J. Besl and N.~D. McKay.
\newblock Method for registration of 3-d shapes.
\newblock {\em IEEE Transactions on Pattern Analysis and Machine Intelligence},
  14(2):239--256, 1992.

\bibitem{boyer2011shrec}
E.~Boyer, A.~Bronstein, M.~Bronstein, B.~Bustos, T.~Darom, R.~Horaud, I.~Hotz,
  Y.~Kelle, J.~Keustermans, A.~Kovnatsky, et~al.
\newblock Shrec 2011: Robust feature detection and description benchmark.
\newblock In {\em Proceedings of the Eurographics Workshop on 3D Object
  Retrieval}, 2011.

\bibitem{brown2007automatic}
M.~Brown and D.~G. Lowe.
\newblock Automatic panoramic image stitching using invariant features.
\newblock {\em International Journal of Computer Vision}, 74(1):59--73, 2007.

\bibitem{buch2014search}
A.~G. Buch, Y.~Yang, N.~Kr{\"u}ger, and H.~G. Petersen.
\newblock In search of inliers: 3d correspondence by local and global voting.
\newblock In {\em Proceedings of the IEEE Conference on Computer Vision and
  Pattern Recognition}, pages 2075--2082. IEEE, 2014.

\bibitem{chen20073d}
H.~Chen and B.~Bhanu.
\newblock 3d free-form object recognition in range images using local surface
  patches.
\newblock {\em Pattern Recognition Letters}, 28(10):1252--1262, 2007.

\bibitem{cho2009feature}
M.~Cho, J.~Lee, and K.~M. Lee.
\newblock Feature correspondence and deformable object matching via
  agglomerative correspondence clustering.
\newblock In {\em Proceedings of the IEEE Conference on Computer Vision and
  Pattern Recognition}, pages 1280--1287. IEEE, 2009.

\bibitem{cho2014finding}
M.~Cho, J.~Sun, O.~Duchenne, and J.~Ponce.
\newblock Finding matches in a haystack: A max-pooling strategy for graph
  matching in the presence of outliers.
\newblock In {\em Proceedings of the IEEE Conference on Computer Vision and
  Pattern Recognition}, pages 2083--2090, 2014.

\bibitem{chum2005matching}
O.~Chum and J.~Matas.
\newblock Matching with prosac-progressive sample consensus.
\newblock In {\em Proceedings of the IEEE Conference on Computer Vision and
  Pattern Recognition}, volume~1, pages 220--226. IEEE, 2005.

\bibitem{enqvist2009optimal}
O.~Enqvist, K.~Josephson, and F.~Kahl.
\newblock Optimal correspondences from pairwise constraints.
\newblock In {\em Proceedings of the IEEE International Conference on Computer
  Vision}, pages 1295--1302. IEEE, 2009.

\bibitem{fischler1981random}
M.~A. Fischler and R.~C. Bolles.
\newblock Random sample consensus: a paradigm for model fitting with
  applications to image analysis and automated cartography.
\newblock {\em Communications of the ACM}, 24(6):381--395, 1981.

\bibitem{guo2016comprehensive}
Y.~Guo, M.~Bennamoun, F.~Sohel, M.~Lu, J.~Wan, and N.~M. Kwok.
\newblock A comprehensive performance evaluation of 3d local feature
  descriptors.
\newblock {\em International Journal of Computer Vision}, 116(1):66--89, 2016.

\bibitem{guo2013rotational}
Y.~Guo, F.~Sohel, M.~Bennamoun, M.~Lu, and J.~Wan.
\newblock Rotational projection statistics for 3d local surface description and
  object recognition.
\newblock {\em International Journal of Computer Vision}, 105(1):63--86, 2013.

\bibitem{guo2014accurate}
Y.~Guo, F.~Sohel, M.~Bennamoun, J.~Wan, and M.~Lu.
\newblock An accurate and robust range image registration algorithm for 3d
  object modeling.
\newblock {\em IEEE Transactions on Multimedia}, 16(5):1377--1390, 2014.

\bibitem{johnson1998surface}
A.~E. Johnson and M.~Hebert.
\newblock Surface matching for object recognition in complex three-dimensional
  scenes.
\newblock {\em Image and Vision Computing}, 16(9):635--651, 1998.

\bibitem{leordeanu2005spectral}
M.~Leordeanu and M.~Hebert.
\newblock A spectral technique for correspondence problems using pairwise
  constraints.
\newblock In {\em Proceedings of the International Conference on Computer
  Vision}, volume~2, pages 1482--1489. IEEE, 2005.

\bibitem{lowe2004distinctive}
D.~G. Lowe.
\newblock Distinctive image features from scale-invariant keypoints.
\newblock {\em International Journal of Computer Vision}, 60(2):91--110, 2004.

\bibitem{mahamud2003segmentation}
S.~Mahamud, L.~R. Williams, K.~K. Thornber, and K.~Xu.
\newblock Segmentation of multiple salient closed contours from real images.
\newblock {\em IEEE Transactions on Pattern Analysis and Machine Intelligence},
  25(4):433--444, 2003.

\bibitem{mian2010repeatability}
A.~Mian, M.~Bennamoun, and R.~Owens.
\newblock On the repeatability and quality of keypoints for local feature-based
  3d object retrieval from cluttered scenes.
\newblock {\em International Journal of Computer Vision}, 89(2-3):348--361,
  2010.

\bibitem{mian2006three}
A.~S. Mian, M.~Bennamoun, and R.~Owens.
\newblock Three-dimensional model-based object recognition and segmentation in
  cluttered scenes.
\newblock {\em IEEE Transactions on Pattern Analysis and Machine Intelligence},
  28(10):1584--1601, 2006.

\bibitem{mian2006novel}
A.~S. Mian, M.~Bennamoun, and R.~A. Owens.
\newblock A novel representation and feature matching algorithm for automatic
  pairwise registration of range images.
\newblock {\em International Journal of Computer Vision}, 66(1):19--40, 2006.

\bibitem{otsu1975threshold}
N.~Otsu.
\newblock A threshold selection method from gray-level histograms.
\newblock {\em Automatica}, 11(285-296):23--27, 1975.

\bibitem{petrelli2015pairwise}
A.~Petrelli and L.~Di~Stefano.
\newblock Pairwise registration by local orientation cues.
\newblock In {\em Computer Graphics Forum}. Wiley Online Library, 2015.

\bibitem{rodola2013scale}
E.~Rodol{\`a}, A.~Albarelli, F.~Bergamasco, and A.~Torsello.
\newblock A scale independent selection process for 3d object recognition in
  cluttered scenes.
\newblock {\em International Journal of Computer Vision}, 102(1-3):129--145,
  2013.

\bibitem{rusu2009fast}
R.~B. Rusu, N.~Blodow, and M.~Beetz.
\newblock Fast point feature histograms (fpfh) for 3d registration.
\newblock In {\em Proceedings of the IEEE International Conference on Robotics
  and Automation}, pages 3212--3217, 2009.

\bibitem{rusu2008aligning}
R.~B. Rusu, N.~Blodow, Z.~C. Marton, and M.~Beetz.
\newblock Aligning point cloud views using persistent feature histograms.
\newblock In {\em Proceedings of the IEEE/RSJ International Conference on
  Intelligent Robots and Systems}, pages 3384--3391, 2008.

\bibitem{rusu20113d}
R.~B. Rusu and S.~Cousins.
\newblock 3d is here: Point cloud library (pcl).
\newblock In {\em Proceedings of the IEEE International Conference on Robotics
  and Automation}, pages 1--4, 2011.

\bibitem{salti2010use}
S.~Salti, F.~Tombari, and L.~Di~Stefano.
\newblock On the use of implicit shape models for recognition of object
  categories in 3d data.
\newblock In {\em Proceedings of the Asian Conference on Computer Vision},
  pages 653--666. Springer, 2010.

\bibitem{shi2000normalized}
J.~Shi and J.~Malik.
\newblock Normalized cuts and image segmentation.
\newblock {\em IEEE Transactions on Pattern Analysis and Machine Intelligence},
  22(8):888--905, 2000.

\bibitem{sipiran2011harris}
I.~Sipiran and B.~Bustos.
\newblock Harris 3d: a robust extension of the harris operator for interest
  point detection on 3d meshes.
\newblock {\em The Visual Computer}, 27(11):963--976, 2011.

\bibitem{tombari2010object}
F.~Tombari and L.~Di~Stefano.
\newblock Object recognition in 3d scenes with occlusions and clutter by hough
  voting.
\newblock In {\em Proceedings of the Fourth Pacific-Rim Symposium on Image and
  Video Technology}, pages 349--355. IEEE, 2010.

\bibitem{tombari2010unique}
F.~Tombari, S.~Salti, and L.~Di~Stefano.
\newblock Unique signatures of histograms for local surface description.
\newblock In {\em Proceedings of the European Conference on Computer Vision},
  pages 356--369. 2010.

\bibitem{tombari2013performance}
F.~Tombari, S.~Salti, and L.~Di~Stefano.
\newblock Performance evaluation of 3d keypoint detectors.
\newblock {\em International Journal of Computer Vision}, 102(1-3):198--220,
  2013.

\bibitem{torr2000mlesac}
P.~H. Torr and A.~Zisserman.
\newblock Mlesac: A new robust estimator with application to estimating image
  geometry.
\newblock {\em Computer Vision and Image Understanding}, 78(1):138--156, 2000.

\bibitem{vc1962method}
H.~P. VC.
\newblock Method and means for recognizing complex patterns, 1962.
\newblock US Patent 3,069,654.

\bibitem{yang2016fast}
J.~Yang, Z.~Cao, and Q.~Zhang.
\newblock A fast and robust local descriptor for 3d point cloud registration.
\newblock {\em Information Sciences}, 346:163--179, 2016.

\end{thebibliography}
}

\end{document}